\documentclass[sn-mathphys,Numbered]{sn-jnl}% Math and Physical Sciences Reference Style
\usepackage{calc}
\usepackage{amssymb}
\usepackage{amstext}
\usepackage{amsthm}
\usepackage{mathtools}
\usepackage{amsmath}
\usepackage{multicol}
\usepackage{plain}
\usepackage{tabularx}
\usepackage{float}
\usepackage{systeme}
\usepackage[T1]{fontenc}
\usepackage{fancyhdr}
\usepackage{algorithm}
\usepackage{algpseudocode}
\usepackage{diagbox}
\usepackage{multirow}
\usepackage{hyperref}
\newcolumntype{M}[1]{>{\centering\arraybackslash}m{#1}}
\usepackage{graphicx}
\usepackage{subfigure}
\usepackage[font=footnotesize,labelfont=bf]{caption}
\usepackage{xparse}

\usepackage{graphicx}%
\usepackage{multirow}%
\usepackage{amsmath,amssymb,amsfonts}%
\usepackage{amsthm}%
\usepackage{mathrsfs}%
\usepackage[title]{appendix}%
\usepackage{xcolor}%
\usepackage{textcomp}%
\usepackage{manyfoot}%
\usepackage{booktabs}%
\usepackage{algorithm}%
\usepackage{algorithmicx}%
\usepackage{algpseudocode}%
\usepackage{listings}%
\usepackage[numbers]{natbib}
\begin{document}

\title{Online hand gesture recognition using Continual Graph Transformers }


\author*[1,2]{Rim Slama}\email{rim.slamasalmi@entpe.}
%Affiliation to adjust
\author[1]{Wael  Rabah}\email{wrabah@cesi.fr}
\author[3]{Hazem Wannous}\email{hazem.wannous@univ-lille.fr}

\affil*[1]{CESI LINEACT, UR 7527, Lyon, 69100, France}

\affil[2]{LICIT-ECO7,
ENTPE, Université Gustave Eiffel, Lyon, 69518, France}

\affil[3]{IMT Nord Europe,
University of Lille, CNRS
UMR 9189 - CRIStAL, Lille, 59000,France}


\abstract{Online continuous action recognition has emerged as a critical research area due to its practical implications in real-world applications, such as human-computer interaction, healthcare, and robotics. Among various modalities, skeleton-based approaches have gained significant popularity, demonstrating their effectiveness in capturing 3D temporal data while ensuring robustness to environmental variations. However, most existing works focus on segment-based recognition, making them unsuitable for real-time, continuous recognition scenarios.
In this paper, we propose a novel online recognition system designed for real-time skeleton sequence streaming. Our approach leverages a hybrid architecture combining Spatial Graph Convolutional Networks (S-GCN) for spatial feature extraction and a Transformer-based Graph Encoder (TGE) for capturing temporal dependencies across frames. Additionally, we introduce a continual learning mechanism to enhance model adaptability to evolving data distributions, ensuring robust recognition in dynamic environments.
We evaluate our method on the SHREC’21 benchmark dataset, demonstrating its superior performance in online hand gesture recognition. Our approach not only achieves state-of-the-art accuracy but also significantly reduces false positive rates, making it a compelling solution for real-time applications. The proposed system can be seamlessly integrated into various domains, including human-robot collaboration and assistive technologies, where natural and intuitive interaction is crucial.}




\maketitle

\section{Introduction}
Hand gesture recognition (HGR) has recently attracted a lot of attention in the human-machine interaction field thanks to the opportunities it offers in different contexts such as health-care, industry and entertainment. 
HGR is the process of detecting and classifying a  sequence of images of the human hand or a sequence of the 2D or 3D coordinates of the joints in the skeleton of the hand into a pre-defined list of gesture categories. There are generally two approaches to HGR: offline and online HGR.
In the offline task, the gesture sequences are segmented. The challenge with offline recognition is to recognize the gesture classes as accurately as possible. As opposed to online recognition, where the sequence is not segmented and can contain one or multiple gestures or can contain no gestures at all. Online recognition brings new challenges, we have to handle the sequential processing of the data stream and perform an on-the-fly classification of the gesture accurately and efficiently. \\

Based on the chosen modality, hand gesture recognition approaches are split into two categories: image-based approaches which employ RGB or RGB-D image sequences and skeleton-based methods which use sequences of 2D or 3D euclidean coordinates of the hand joints \cite{hand-gesture-survey}.
The rapid development of low-cost depth sensors such as Microsoft Kinect, leap motion and Intel RealSense, coupled with quick advances in hand pose estimation research, has allowed the capture of hand skeleton data with high precision. \\

Various Skeleton-based approaches \cite{lai2018cnn+,du2015hierarchical,shin2020skeleton,li2017skeleton1,li2017skeleton2} were proposed in recent years, which use traditional Convolutional Neural Networks (CNN), Recurrent Neural Networks (RNN) or Long Short Term Memory (LSTM). Most of these methods treat the hand gesture sequence as a time series and mainly focus on capturing the temporal features by using RNN  based models, which are not very efficient and fail to retain long-term dependencies and to exploit the natural connectivity among the joints. In this project, we have decided to focus on 3D skeleton data modality. Skeleton-based hand gesture recognition is a challenging task that sparked a lot of attention in recent years, especially with the rise of Graph Neural Networks. \\

In a first step, we evaluate the performance of our method on three offline benchmarks: the SHREC'17 Track dataset, Briareo dataset and the First Person Hand Action dataset. The experiments show the efficiency of our approach, which achieves similar performance or outperforms the state of the art.
In a second step, we integrate our model into an online method that performs real-time hand gesture recognition. We test this online method on the SHREC'21 track dataset \cite{shrec21}. \\

\section{Related works}
Skeleton-based HGR has become an active research area in recent years, and it has been studied extensively, especially with the rise of deep learning. This led to the development of many advanced skeleton-based approaches \cite{akremi2022spd,caputo,dhg,sem-mem-wal,res-tcn,HPEV+HMM+FRPV,st-ts-hgr-net,huang-Riemannian,huang-grassman}. \\
We will start this state-of-the-art section by presenting some background behind 3D skeleton data collection, and we will list all publicly available 3D Hand Gesture Recognition datasets that feature skeletal data, and then we will present the most recent literature related to the task.

\subsection{Offline hand gesture recognition methods}
In this section, we only focus on 3D skeleton based offline methods. Skeleton based HGR methods are usually divided into 2 groups : hand-crafted features based methods and deep learning based methods. 
\subsubsection{Handcrafted features methods}
Handcrafted features are properties that are derived or computed from the original data, they serve the purpose of enriching the feature vectors and extract more information from the original feature vector. We can take the example of computing the velocity between a joint $i$ in a frame $t$ and its parallel joints $i$ from other frames in the sequence, the velocity in this example is considered a handcrafted feature, and it should provide information about the temporal evolution of each joint. \\
Handcrafted features methods tend to encode the 3D skeleton feature vectors into other feature descriptors,  this list includes position, motion, velocity and orientation descriptors and this led to researchers exploiting handcrafted features for HGR, as it turned out that these features provide a good description of the hand movement. In most cases, they use these features as input to a supervised learning classifier like Support vector machines (SVM) or Random Forests in \cite{ohn2013joint, de2019heterogeneous, de2016skeleton}. 
\subsubsection{Deep learning methods}
Deep learning methods use Convolutional neural networks (CNN) and Recurrent neural networks (RNN) to encode the skeleton data into spatial temporal feature vectors. However, these networks do not exploit the adjacency between the joints or the correlation between the hand joints between different frames \cite{chen2017motion,nunez2018convolutional,res-tcn}.\\

In this work, we use GCNs \cite{gcn}, attention and transformers \cite{transformer} so for the rest of this section, we will only focus on recent skeleton based gesture and action recognition state-of-the-art methods that use these models.\\
\paragraph{GCN based approaches}
One of the first approaches that use GCNs on skeleton data was spatial temporal graph convolution networks (ST-GCN), proposed by \textbf{Yan et al.} \cite{st-gcn}, in which they construct a spatio-temporal graph from a 3D skeleton body. Spatial graph convolution and temporal convolution layers were introduced to extract the adequate features from the body graph sequence. Some other approaches developed new architectures inspired by ST-GCN. 
AS-GCN \cite{as-gcn} introduced new modules to the ST-GCN architecture that capture actional and structural relationships. This helps them overcome ST-GCN's disregard for hidden action-specific joint correlations.  
Non-local graph convolutions \cite{non-local-gcn} proposed to learn a unique individual graph for each sequence. Focusing on all joints, they decide whether there should be connections between pairs of joints or not. 2S-AGCN \cite{shi2019two} built a 2 stream architecture to model both the skeleton data and second-order information such as the direction and length of the bones. They used Adaptive GCN (AGCN) \cite{li2018adaptive}, which learns 2 adjacency matrices individually for each sequence and uniformly shared between all the sequences. The same authors later proposed MS-AAGCN \cite{shi2020skeleton} that improves on their previous architecture \cite{li2018adaptive} by modeling a third stream called the motion stream. AAGCN was proposed, which further enhances on AGCN with a spatio-temporal attention module, enabling the learned model to pay more attention to important joints, frames and features. \\
\paragraph{Attention and Transformer based approaches}
Transformers are sequence models introduced primarily in NLP, which perform better feature extraction than recurrent models thanks to the self-attention mechanism. The most recent and notable related works include STA-GCN \cite{sta-gcn} which used spatial and temporal self-attention modules to learn trainable adjacency matrices. In STA-RES-TCN \cite{res-tcn}, spatio-temporal attention was used to enhance residual temporal convolutional networks. The use of the attention mechanism enables the network to concentrate on the important frames and features and eliminate the unimportant ones that frequently add extra noise. In Attention Enhanced Graph Convolutional LSTM Network (AGCLSTM) \cite{si2019attention}, they extract three types of features. They capture spatial connections and temporal dynamics, and in addition to that they study the co-occurrence link between spatial and temporal domains also the attention mechanism is used to produce more informative features of important joints. DG-STA \cite{dg-sta} proposed to leverage the attention mechanism to construct dynamic temporal and spatial graphs by automatically learning the node features and edges. ST-TR \cite{st-tr} proposed a Spatial and temporal Self-Attention modules used to understand intra-frame interactions between different body parts and interpret hidden inter-frame correlations. To solve a similar problem, full body gesture recognition, the authors of "Skeleton-Based Gesture Recognition Using Several Fully Connected Layers with Path Signature Features and Temporal Transformer Module"  \cite{li2019skeleton} used path signature features. This feature is used as a trajectory descriptor for each single joint and joint pair by encoding the spatial and temporal paths that this joint follows through the sequence, it serves as a good indication on how the data travels through the sequence. They use fully connected layers and a Temporal Transformer Module for feature extraction. \\

\subsection{Online hand gesture recognition methods}
Due to the lack of online HGR in recent, there aren't many GCN and attention based approaches in the literature. In this section, we will discuss recent online skeleton-based HGR methods in general. \\
Recently, Many new online methods were proposed with the creation of the SHREC'21 dataset \cite{shrec21}. In the contest, 4 methods were proposed by 4 research groups, we will briefly explain each one of their approaches: \\

Group 1 \cite{shrec21} proposed a transformer based network as their recognition model. For online recognition, they use a sliding window approach coupled with a Finite State Machine to detect when gestures start. The FSM starts at state S1, it uses a buffer to store 10 past frames, and each frame is classified. If even one frame is classified as a gesture, then the state of the FSM is increased to S2 and a check on the beginning of the gesture is performed. The system checks for 10 consecutive windows, if not enough gestures are found, then the FSM empties the buffer and abandons this gesture and resets to the initial state S1. If enough gestures are detected, then the state is increased  to S3. The FSM attempts to detect the end of the gesture and checks if a window does not contain any gestures, then the state is increased to S4. In this state, the FSM checks in the next 25 consecutive windows. if at least one gesture is detected, the state is decreased to S3. Otherwise, if no gesture is detected, the FSM resets to the initial state S1 and the whole process is restarted. \\
Group 2 \cite{shrec21} proposed an image based approach. They project the 3D skeletons on the xy plane to create a sequence of 2D images, and then  they use ResNet-50 \cite{he2016deep} which is a Convolutional Neural Network (CNN) based model as their classification model. To provide temporal information to the network, the recent history of the gesture is traced on the image to finally produce only one image to represent a sequence of 3D skeletons. There are several disadvantages with this method, the lack of a true temporal dimension means that they can't exploit the temporal correlations between the frames. They proposed a second model to solve this issue, by using the ResNet-3D network \cite{hara2018can}, they are able to exploit the temporal dimension of this data. But another disadvantage is that they ignored the z-axis in the data, which can affect the recognition rate for some gestures that exploit all 3 dimensions. Their online recognition is not really continuous, they perform the recognition every 300 frames, which is a very long window that can include multiple gestures, so some short gestures can be ignored. This explains why their methods produce more false positives than the other 3 groups. \\
Group 3 \cite{shrec21} proposed uDeepGRU, a model based on Recurrent Neural Networks (RNNs), that uses a network of Gated Recurrent Units (GRUs). They classify the frames sequentially as they are fed to the network and outputs a predicted label for every frame. They obtained decent results, despite that the network doesn't explicitly exploit the spatial features of the 3D skeletons.\\
Group 4 \cite{shrec21} used st-gcn \cite{st-gcn} as their classification model. For online recognition, they used a sliding window approach, and they use an energy-based pre-segmentation module that calculates the amount of energy accumulated in a window W, to determine if the classification should be performed on W or to ignore it and consider it a "non-gesture" window. This pre-segmentation module helped them reduce the number of false positives exponentially, and they performed the best overall out of all 4 groups.\\

Outside of the SHREC'21 contest, the STRONGER method \cite{emporio2021stronger} was evaluated on the SHREC'21 dataset. They proposed a network of simple 1D CNNs, each coordinate out of (x, y, z) is processed by one 1D CNN. The values of each coordinate from all the frames are concatenated to form a vector that represents the temporal evolution in each dimension, and are then fed to the 1D CNN to extract features. All extracted feature vectors are then concatenated and fed to the classifier to predict the label of the sequence. For online recognition, they also use a sliding window approach and a threshold filtering system to reduce false positives. They learn some thresholds during the training phase, which they use later in the online recognition phase to filter the windows with a low confidence score and consider them as "non-gesture" windows. 

\section{CoSTrGCN:  Continual Transformers to our approach STrGCN}

Our approach STrGCN performs well on the task of offline HGR. We evaluated STrGCN on the SHREC21 online benchmark \cite{shrec21}, by introducing a sliding window approach and a prediction
filtering system based on probability thresholds, and the model performed well on this setup. We attempt to further improve this approach by updating the architecture of our model by introducing a memory mechanism, which was inspired by the work of Hedegaard et al. in continual inference \cite{hedegaard2022continualinference} and continual transformers \cite{hedegaard2022continual}

\subsection{Overview of the approach}
\begin{figure*}[ht!]
  \makebox[1.\textwidth][c]{\includegraphics[width=.87\paperwidth]{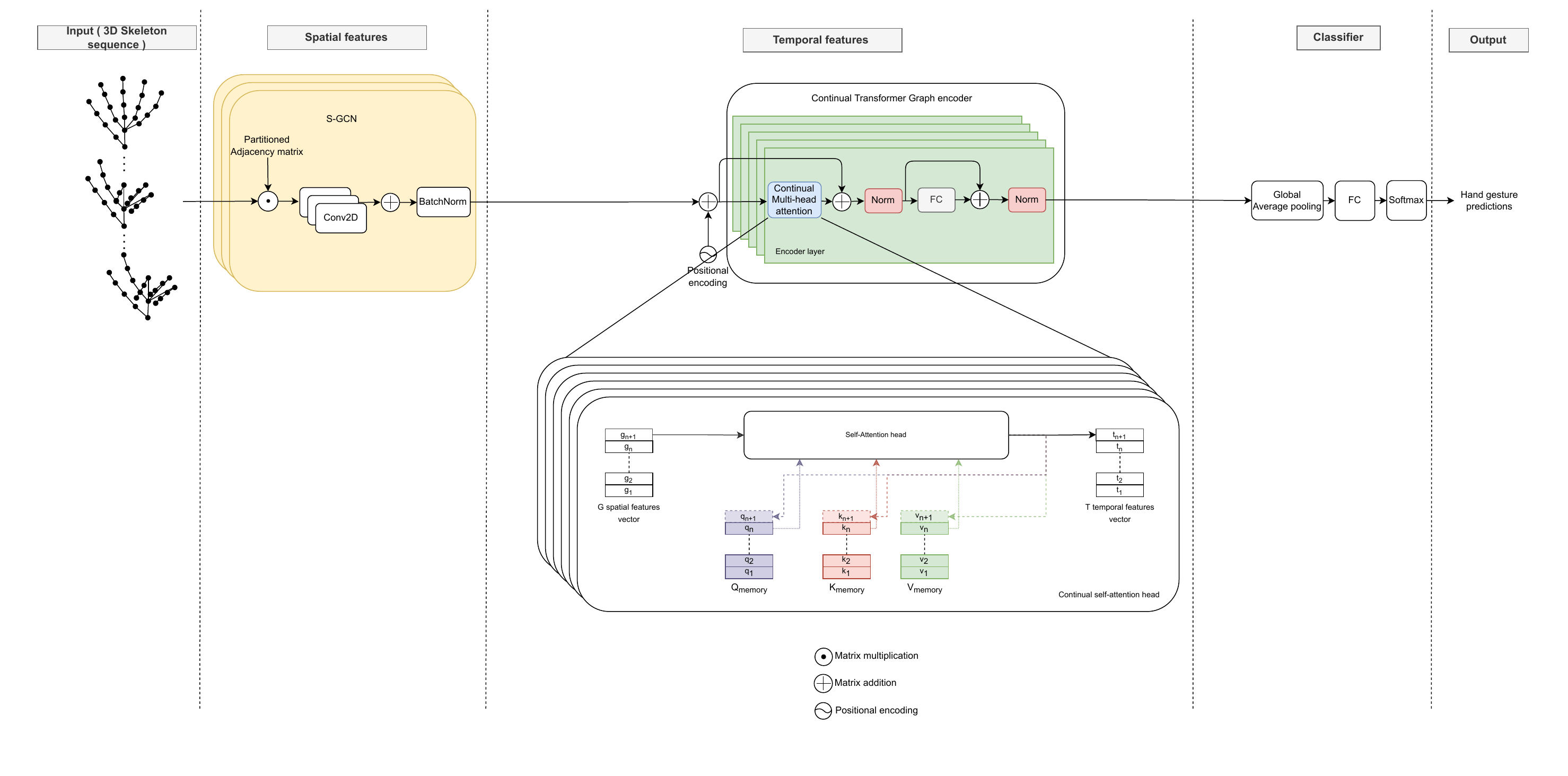}}%
  \ \caption{Overview of the CoST-GCN framework for 3D skeleton-based action recognition, integrating spatial and temporal feature extraction. The architecture consists of a spatial GCN module for capturing spatial relationships, followed by a Contextual Transformer Graph Encoder for temporal dependencies. The output is processed through a classifier to predict hand gesture actions. }
  \label{CoSTrGCN}
\end{figure*}
Continual-inference and continual transformers are used for stream-processing, where there's an endless stream of continuous data incoming from one or multiple detection devices. This creates the need of a more robust architecture that can handle enormous amounts of data. \\
We aim to apply these concepts on our study case, We study the application of continual inference on out Transformer Graph Encoder module to allow for online processing of 3D hand skeleton data. This is achieved by processing a single hand skeleton at each time-step to compute the query, key and value vectors of this frame. The goal is to produce and handle a continuous stream of query, key, and value vectors in the form of $d\textsubscript{model}$-dimensional tokens, and in order to reduce redundant computations, we store the n previous computations of $q$,$v$ and $k$, and at each time step, we only compute the $q$,$v$ and $k$ vectors of the $nth$ time step. the outputs of the $TGE$ for each step is computed based on the n previous tokens in real-time despite that we process only one hand skeleton at each time-step.

%------------------------------------------------------------------------- 
In this section, we present the architecture of the proposed network. First, we give an overview of the whole process. Then, each module is illustrated separately.

%\begin{figure*}[ht!]
%  \makebox[1.\textwidth][c]{\includegraphics[width=.87\paperwidth]{images/Full_archi.pdf}}%
%  \ \caption{This figure shows an overview of our approach. }
%  \label{FullArchi}
%\end{figure*}

In our method, we take advantage of two models: a Spatial Graph Convolutional Network (S-GCN) is used for spatial information extraction from graphs, coupled with a Transformer Graph Encoder (TGE) for capturing temporal features in sequences. Fig. \ref{CoSTrGCN} describes different units of our proposed architecture for skeleton-based hand gesture recognition. The challenge with this architecture was to adapt the Transformer to handle graph sequences instead of word sequences. \\ %handling sequences of different lengths without losing performance, unlike recurrent models that witness a drop in performance as the sequences get longer. 
Having a sequence of 3D hand skeletons, we use an adjacency matrix to construct a graph sequence. First, in the spatial domain, S-GCN is used to extract hand features at each frame taking advantage of the natural graph structure of the hand skeleton. Then, in a temporal domain and respecting the graph structure of the spatial features, a transformer graph encoder is proposed to extract inter-frame relevant features. Finally, a global pooling operation is used to aggregate the graph into a representation that can be interpreted by our classifier.
\subsection{Data modellisation}
The global input to this model is a 3D skeleton sequence represented in the form of a spatial temporal graph. This graph is showcased in figure. \ref{CoSTrGCN}
\subsection{Problem formulation}
The proposed gesture recognition approach can be defined as a function denoted by $GT$:
\begin{equation}
\begin{split}
    & GT: \mathbb{R}^{\gamma{*\lambda{*\phi}}} \rightarrow \mathbb{R}^C \\
    & GT \mapsto Y(T(G(S)+PE))    
\end{split}
\end{equation}

This function predicts a probability distribution over $C$ gesture classes from a sequence ${S \in \mathbb{R}^{\gamma{*\lambda{*\phi}}}}$ sampled out of the input set of 3D skeleton sequences. We denote, by ${\gamma}$ the sequence length, $\lambda$ the number of hand skeleton nodes constructing the graph and $\phi$ the number of features per node.

$G$ is the function of the S-GCN module, $PE$ stands for the positional encoding and $T$ represents the TGE module and $Y$ represents the classifier.
$GT$ can be decomposed into three operations: \\
{\bf The first operation} $G$ corresponds to the {\bf S-GCN} for spatial features extraction from a hand skeleton:
\begin{equation}
\begin{split}
    & G: \mathbb{R}^{{\gamma}*\lambda*\phi} \rightarrow            \mathbb{R}^{{\gamma}*\lambda*{d\textsubscript{model}}} \\
    & \hat{g}=G(S) 
\end{split}
\end{equation}

${d\textsubscript{model}}$ is the embedding size of each graph node. An embedding is a numerical vector representation of a complex structured data object. It is calculated such that 2 similar objects would have 2 similar embedding vectors. In our case, $\hat{g}$ is a collection of embedding vectors of each node of the graph. \\ 
{\bf The second operation} $T$ corresponds to the TGE module for temporal feature extraction from inter-frame hand joints:

\begin{equation}
\begin{split}
    & T: \mathbb{R}^{{\gamma}*\lambda*{d\textsubscript{model}}}
    \rightarrow \mathbb{R}^{{\gamma}*\lambda*{d\textsubscript{model}} }\\
    & \hat{t}=T(\hat{g}+PE)
\end{split}
\end{equation}
$PE$ is the Positional Encoding vector. The sequence in its current state doesn't carry information about the position of each frame in the sequence. Thus, we add a $PE$ vector to $\hat{g}$, which is a vector that encodes the position of each frame in the sequence. You can refer to Vaswani et al. \cite{transformer} for more information about the PE operation. \\
%{\bf The third operation} $P$ is the global pooling which transforms a sequence of temporal graph features into an aggregated feature map of size ${d\textsubscript{model}}$:
\begin{equation}
f=P(t) \ with \ P: \mathbb{R}^{{\gamma}*\lambda*{d\textsubscript{model}}} \rightarrow \mathbb{R}^{d\textsubscript{model}}
\end{equation}
.
{\bf The third operation} $Y$ is the classifier. First, we apply a global pooling operation  $P$ that transforms a sequence of temporal graph features into an aggregated feature map $\hat{f} \in \mathbb{R}^{d\textsubscript{model}}$. This pooling operation consists in average pooling each dimension of the sequence separately, average pooling on the graph $\lambda$ dimension of the vertices/nodes in a first step and then average pooling on the $\gamma$ dimension of the frames in a second step, we get a vector of size $d\textsubscript{model}$. 
Then a $FC$ layer maps the aggregated temporal features into the $C$ potential gesture classes respecting the following formula:
\begin{equation}
\begin{split}
    & Y: \mathbb{R}^{d\textsubscript{model}} \rightarrow \mathbb{R}^C \\
    & \hat{y}=Y(\hat{f})
\end{split}
\end{equation}
We denote by $\hat{y}$ the probability distribution over $C$ gesture classes.
%------------------------------------------------------------------------- 
\subsection{Spatial Graph Convolutional Networks (S-GCN) } \label{s-gcn}
%------------------------------------------------------------------------- 
\begin{figure*}[ht!]
  \makebox[1.\textwidth][c]{\includegraphics[width=.87\paperwidth]{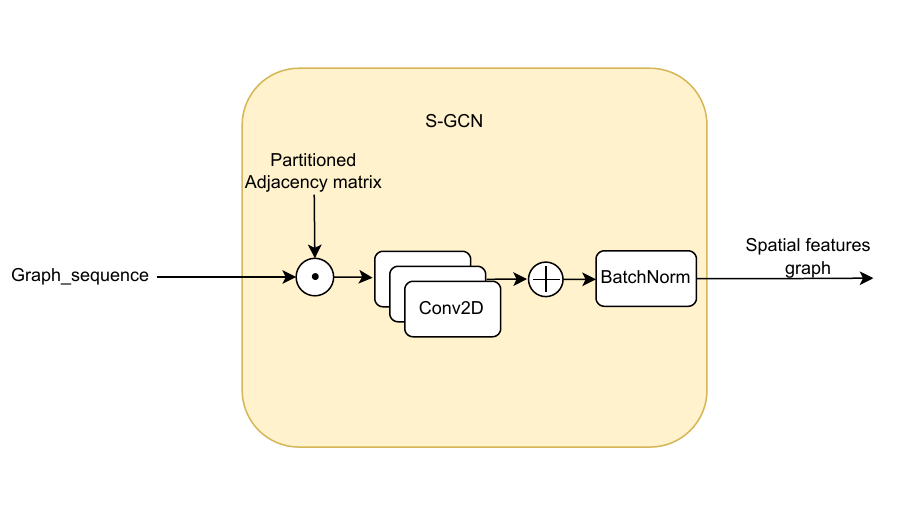}}%
  \ \caption{ S-GCN unit }
  \label{sgcn}
\end{figure*}
The spatial graph convolution operation shown in Fig \ref{sgcn} is a weighted average aggregating the features of each node with those of all of its neighbors to produce a new feature vector for each node. This vector contains information about the current node, its neighbors in the same frame, and the importance degree of the connections between them in the hand. We can stack these S-GCN units in a row so that the input of each layer would be the output of the previous one. Stacking this S-GCN unit would allow us to extract more powerful and more global features about the hand.

In order to construct our spatial graph, we need to build an adjacency matrix that represents the connections between different joints in the hand skeleton. We formulate the final adjacency matrix A\textsubscript{k} as follows:
\begin{equation}
A\textsubscript{k} = D\textsubscript{k}^{-1/2}.(\tilde{A\textsubscript{k}} + I).D\textsubscript{k}^{-1/2}, D\textsubscript{ij}=\sum_{k}^{K\textsubscript{a}}(\tilde{A\textsubscript{k}}^{ij} + I\textsubscript{ij})
\end{equation}
$\tilde{A\textsubscript{k}}$ is the adjacency matrix of the fixed undirected graph representing the connections between the hand joints. If we apply the graph convolution on $\tilde{A\textsubscript{k}}$, the result will not contain the features of the node itself. We add $I$, an identity matrix, to represent the self-connections of the nodes. As ($\tilde{A\textsubscript{k}} +I$) is a binary and not normalized matrix. We multiply it by $D\textsubscript{k}^{-1/2}$ (the inverse of the degree matrix of the graph) on both sides to normalize it. Then, we use the following formula to compute the graph convolution:
\begin{equation}
 G(S)=\sum_{k}^{K\textsubscript{a}}(S.A\textsubscript{k}).W\textsubscript{k}, \ S \in \mathbb{R}^{{\gamma}*\lambda*\phi }
\end{equation}
Where $S$ is a 3D skeleton sequence and $K\textsubscript{a}$ is the kernel size on the spatial dimension, which also matches the number of adjacency matrices. The number of adjacency matrices depends on the used partitioning strategy, which we will explain below. $W\textsubscript{k}$ is a trainable weight matrix and is shared between all graphs to capture common properties. \\ For each kernel, $S.A\textsubscript{k}$ calculates the weighted average of the features of each node with its neighboring nodes, which is then multiplied by the weights' matrix $(S.A\textsubscript{k}).W\textsubscript{k}$. The features calculated by all the kernels are then summed up to form one feature vector per node.
 In this module, we are inspired by the graph convolution from ST-GCN \cite{st-gcn}, which uses a similar Graph Convolution formulation to the one proposed by Kipf et al. \cite{gcn}. 
 We use the partitioning and edge importance techniques to extract more informative features about the node's neighbors and the edges connecting the nodes.
 
 \subsubsection*{ Partitioning}
 \label{partitioning}
 Inspired by Yan et al. \cite{st-gcn}. We adopt the partitioning technique that consists in dividing the original graph into multiple equal sized subsets. We use it to partition the neighbor set of each node into multiple partitions. In their paper, they suggested three partitioning strategies: 
\begin{itemize}
    \item uni-labeling :  With this strategy, we have two subsets: the first subset only contains the root node. The second subset represents the neighbors set for each node. It consists in setting the subset according to the natural connections between the nodes. The graph convolution would be equivalent to multiplying the feature vector of each node with the average feature vector of all its neighboring nodes.
        \begin{figure}[htbp]
    \centerline{\includegraphics[width=.5\columnwidth]{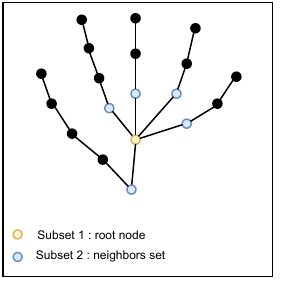}}
     \caption{The "Uni-labeling" partitioning strategy".}
    \label{UniLabelingPartitioning}
    \end{figure}
   % \FloatBarrier
    \item distance partitioning : this strategy consists in setting the partitions according to the node's hop distance in comparison to the root node. The hop distance is the number of edges that separate two nodes. A subset of neighbors with a hop distance of 0 is the root node itself, a subset with a hop distance of 1 will include the root node and its direct neighbors and a subset with a hop distance of 2 will include the root node, its direct neighbors and the neighbors of its neighbors.
    \begin{figure}[htbp]
    \centerline{\includegraphics[width=.5\columnwidth]{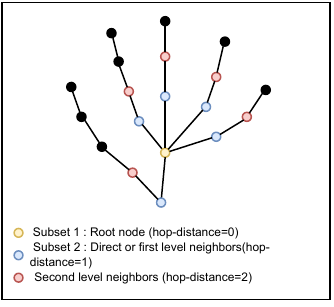}}
     \caption{The "distance" partitioning strategy".}
    \label{DistancePartitioning}
    \end{figure}
    %\FloatBarrier
    \item spatial configuration partitioning : Yan et al. designed a new strategy in  \cite{st-gcn}. This strategy consists in dividing the neighbor set into 3 subsets based on their distance to the gravity center of the skeleton, they considered the gravity center of the skeleton to be the average of all of its node. The subsets are arranged as follows : one subset will contain the root node itself, the centripetal subset will contain the nodes that are closer than the root node to the center of gravity, the centrifugal subset will contain the nodes that have longer distances to the center of gravity than the root node.
    \begin{figure}[htbp]
    \centerline{\includegraphics[width=.5\columnwidth]{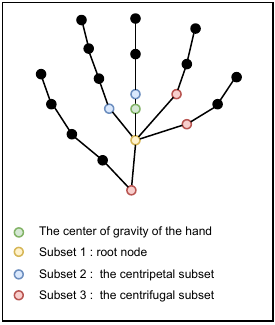}}
     \caption{The "spatial configuration" partitioning strategy".}
    \label{SpatialConfigPartitioning}
    \end{figure}
    %\FloatBarrier
\end{itemize}

We have decided to work with the distance partitioning, which we adapt to the 3D hand skeleton graph (see Fig. \ref{DistancePartitioning}). We chose this strategy because it allows us to capture both local and global features of the hand skeleton.

\subsubsection*{ Edge importance}
This operation is useful when a node contributes to the motion of several surrounding nodes, but these contributions are not equally significant. This mechanism adds a learnable mask for each convolution layer that learns the contribution of each edge to the movement of different parts of the hand. Then, node features are scaled according to the contribution of their edges to their neighboring nodes. 
\begin{figure}[htbp]
\centerline{\includegraphics[width=.5\columnwidth]{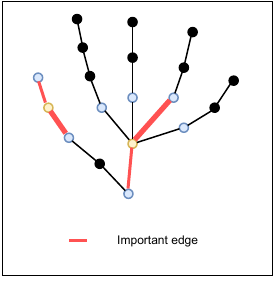}}
 \caption{Edge importance}
\label{EdgeImportance}
\medskip
This figure of the hand skeleton shows the impact of edge importance. The bolder the edge, the more it contributes to the movement of the root node.
\end{figure}
%\FloatBarrier
\subsection{ Transformer Graph Encoder (TGE)}
We propose a Transformer Graph Encoder (TGE) module, which learns the inter-frame correlations locally between joints. These local features are then used to learn the global motion of the hand. As shown in Fig. \ref{CoSTrGCN}, this module is composed of multiple identical encoder layers stacked together, with each one feeding its output to the next. Each encoder layer applies two operations: {\bf multi-head attention} and a simple {\bf fully connected feed-forward network}. Following He et al. \cite{skip-conn} and Ba et al. \cite{layer-norm}, we respectively employ a skip connection around each of the two operations followed by a layer normalization step. This operation improves deeper models, making them easier and faster to optimize without sacrificing their performance. % The output of each operation is $LayerNorm(x + Sublayer(x))$, where, $Sublayer(x)$ is either the multi-head attention or a fully connected neural network that serves the function of extracting features from the output of the multi-head attention.

This TGE module was inspired from the encoder block introduced in the transformer paper \cite{transformer}.  The original Transformer Encoder (TE) handles sequences of words, which are represented as numerical vectors. However, our hand's spatial features are represented by a graph. Thus, we have to adapt `TE' to work on a non-linear graph structure. We alter the self-attention mechanism so that it is computed between the individual nodes of a graph sequence. 
%Besides, the self-attention mechanism is changed in a way that dense layers are replaced with Conv2D layers.   \\

\begin{figure*}[htbp!]
    
        \centering
        
        \includegraphics[width=0.9\textwidth]{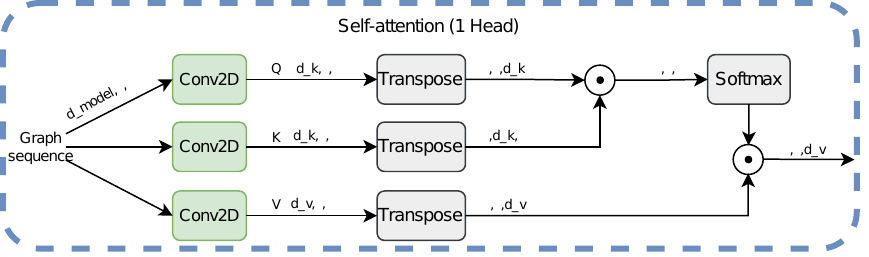} % second figure itself
        \caption{ Self-Attention head.}
        \label{attentionhead}
\end{figure*}

{\bf Multi-head attention}:
\begin{figure*}[htbp!]
        \centering
        \includegraphics[width=0.6\textwidth]{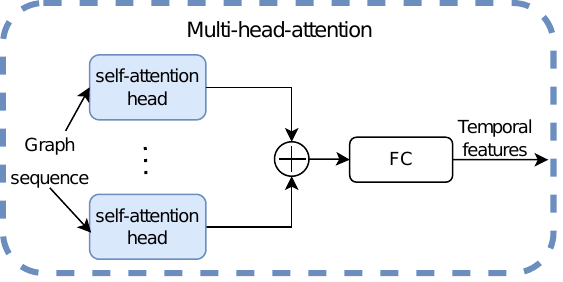} % first figure itself
        \caption{Multi-head self-attention mechanism.}
        \label{multiheadattention}
\end{figure*}
Multi-Head Attention is a self-attention layer that models the inter-frame relationship between the hand joints.
An example of the used multi-head attention module is represented in Fig. \ref{multiheadattention}. To formulate it in the following, we set the number of heads to four:
\begin{equation}
mhAtt(x)=FC(Att_{1} \bigoplus Att_{2} \bigoplus Att_{3} \bigoplus Att_{4}  ) 
\end{equation}
$\bigoplus$ is the concatenation operator. $Att\textsubscript{i}$ is a self-attention head. Each head is initialized differently. In theory, this should allow each head to extract different features from the graph sequence. $FC$ is a fully connected layer that projects the concatenation of our attention heads into a 512-dimensional feature space. Attention is computed between each joint $i \in [1..\lambda]$ at frame $t$, and its corresponding joint in all other frames $t \in [1..\gamma]$. Self-attention  $Att\textsubscript{i}$ is denoted by this formula and represented in Fig. \ref{attentionhead}:
\begin{equation}
\begin{split}
    & Att\textsubscript{i}: \mathbb{R}^{{\gamma}*\lambda*{d\textsubscript{v}}} \rightarrow  \mathbb{R}^{{\gamma}*\lambda*{d\textsubscript{v}}} \\
    & Att\textsubscript{i}(x)=softmax(Q\textsubscript{i}.K\textsubscript{i}/\sqrt{d\textsubscript{k}}).V\textsubscript{i}
\end{split}
\label{att_eq}
\end{equation}

where $x=\hat{g}+PE$ is the spatial features, $W_{Qi}$, $W_{Ki}$ and  $W_{Vi}$ are trainable weight matrices. $Q_{i}$, $K_{i}$ and $V_{i}$ are independent projection matrices of the joints spatial feature vectors and are computed as follows: $Q_{i}=xW_{Qi} \in \mathbb{R}^{\lambda*{\gamma}*{d_{k}}}$, $K_{i}=xW_{Ki} \in \mathbb{R}^{\lambda*{\gamma}*{d_{k}}}$, $V_{i}=xW_{Vi} \in \mathbb{R}{^\lambda*{\gamma}*{d_{v}}}$.
%are independent linear projections of the feature vector of each joint into a 32-d feature space  \\
$d_{v}=32$ is the feature size of the matrix $V_{i}$. $d_{k}=32$ is the feature size of matrices $K_{i}$ and $Q_{i}$. It is also used as a scaling factor, $Q_{i}.K_{i}$ is multiplied by $1/\sqrt{d_{k}}$ to scale large dot product values. 

The operation in equation \ref{att_eq} is called the scaled dot product, which is calculated locally between the joints.
An example of the self-attention head is shown in Fig. \ref{attentionhead}.
The dot product between the $Q_{i}$ and $K_{i}$ matrices is calculated and then a softmax is applied, resulting in a matrix of scores between 0 and 1. The scores are then multiplied by $V_{i}$ to produce the new feature vectors of the joints.
The purpose of this operation is to study the movements of each joint separately and to measure the contribution of its local movements to the global movements of the hand.

%------------------------------------------------------------------------- 

\subsection{Online action training and recognition Formulation}
We base our new continual $TGE$ module on the "CONTINUAL SINGLE-OUTPUT SCALED DOT-PRODUCT ATTENTION" from \cite{hedegaard2022continual}. With this approach to the "SCALED DOT-PRODUCT ATTENTION", we use the traditional formulas adapted to receive the input from one single time-step, then $q\textsubscript{t} k\textsubscript{t} v\textsubscript{t}$ are calculated and cashed for future time-steps. The attention of a single query token $"q\textsubscript{t}"$ is calculated, and one new output is produced, the "CONTINUAL SINGLE-OUTPUT SCALED DOT-PRODUCT ATTENTION" is formulated as follows: \\

\begin{equation}
\begin{split}
    & CoSoAtt_{t}: \mathbb{R}^{1 \times \lambda \times d_{v}} \rightarrow  \mathbb{R}^{1 \times \lambda \times d_{v}} \\
    & CoSoAtt_{t}(q_{t}, k_{t}, v_{t}) = \text{softmax} \left( \frac{q_{t} \cdot (k_{t} \, || \, K_{\text{Mem}})}{\sqrt{d_{k}}} \right) \cdot (v_{t} \, || \, V_{\text{Mem}})
\end{split}
\label{att_eq}
\end{equation}

\section{Experimental results}

Online hand gesture recognition brings new challenges, among which are: distinguishing between parts of the sequence that contain gestures and the ones with no gestures, real-time processing and classification. There are also optimization challenges, the recognition has to be done as efficiently as possible by reducing the delay between the gesture and the classification. For this task, we test our online approach on one benchmark, SHREC 2021 \cite{shrec21}, which we will explain in more details in the data understanding and preparation section.

\subsection{Model parameters}
We conduct our experiments using PyTorch on an NVIDIA Quatro RTX 6000. Adam was used as an optimizer and Cross-entropy as the loss function. For our hyperparameters, a batch size of 32 was chosen for training. The initial learning rate was set to 1e-3, reduced by a factor of 2 if the learning stagnates and the loss doesn't improve in the next 5 epochs. The training stops if the validation accuracy doesn't improve in the next 25 epochs. We set the number of encoders to 6, the number of heads for the multi-head attention to 8 and the value of $d\textsubscript{model}$ to $128$. Taking into account the context of each, a fixed number of frames $\gamma$ was chosen for each dataset: 30 for SHREC'17, 40 for Briareo and 100 for FPHA dataset. To avoid overfitting and improve our model performance, we choose to augment data by random moving. Introduced in  \cite{st-gcn}, it is a form of random affine transformations applied to the sequence to generate an effect similar to moving the angle of the view point of a camera on playback.
We also augment data by noise augmentation, which consists in adding very small values of random noise to the input sequence. 
This also prevents overfitting and allows the model to generalize.
 L1 \cite{l1} and L2 \cite{l2} Regularization techniques and dropout with a rate of 0.3 are used. We initialize the weights of our model by the Xavier initialization \cite{glorot2010understanding} such that the variance of the activation functions is the same across every layer. This initialization contributes to a smoother optimization and prevents exploding or vanishing gradients.

\subsection{Datasets}
In our data collection step, we researched for online hand gesture recognition specific datasets. We found SHREC 2021: Track on Skeleton-based Hand Gesture
Recognition in the Wild \cite{shrec21}, which is a contest focused on online hand gesture recognition. They created a dataset for the contest, which is the only publicly available benchmark specifically designed for the task of online detection and classification of 3D hand gestures. The dataset contains complex and heterogeneous gestures of various types and lengths. \\ 
Unlike the datasets used in the previous step, the gestures in this dataset are not segmented. We used this dataset to train our STr-GCN model, and to evaluate it in an online setting.\\
The SHREC'21 dataset features skeleton sequences composed of 20 joints, 180 sequences were recorded in total by 5 subjects. Each captured sequence contains either 3, 4 or 5 gestures seperated by fragments of the sequence not containing any gestures. The dataset is divided into a training set composed of 108 sequences containing 24 occurrences of each gesture and a test set composed of 72 sequences containing 16 occurrences of each gesture. This dataset has a large and diverse gesture dictionary composed of 18 gesture classes divided into 3 categories :
\begin{figure}[!htbp]
\begin{center}
\includegraphics[width=\columnwidth]{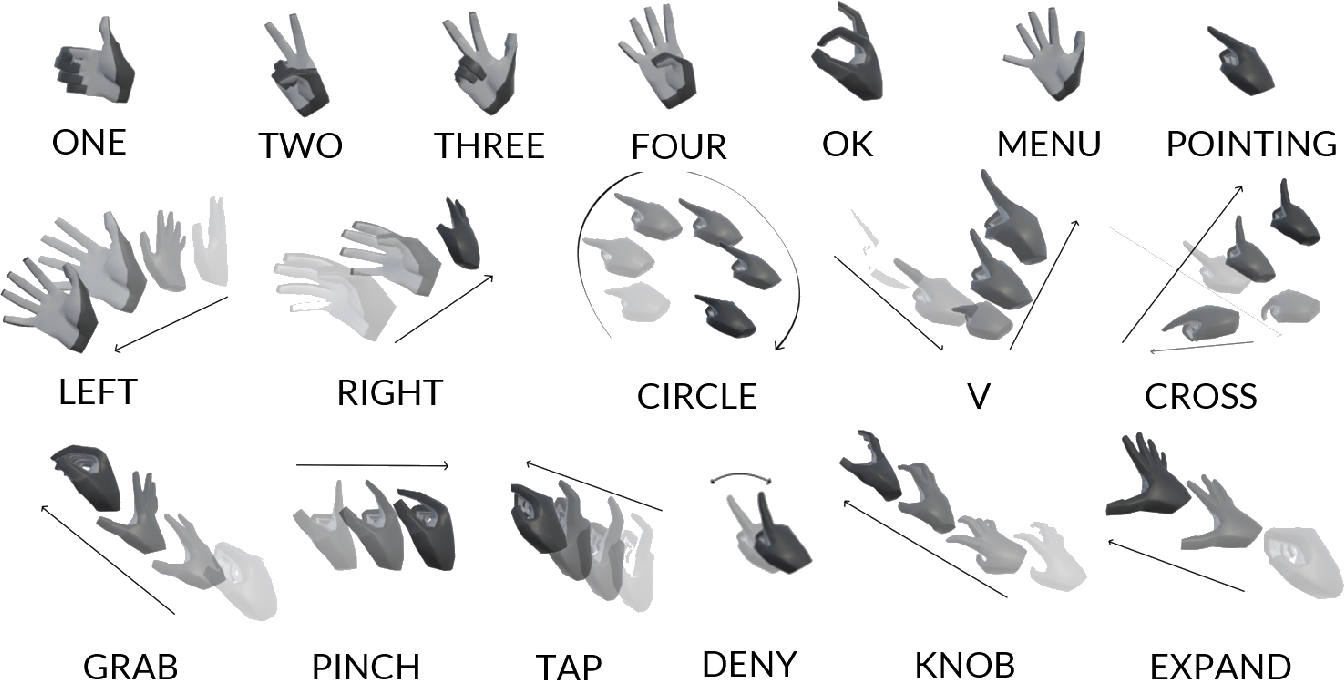}
\caption{ Gesture templates for the SHREC'21 dataset \cite{shrec21}}
\label{shrec21_gesture_classes}
\end{center}
\end{figure}
\begin{itemize}
    \item {\bf static gestures} : characterized by the hand skeleton kept stationary for at least one second. This category includes 7 gesture classes : One, Two, Three, Four, OK, Menu and Pointing.
    \item {\bf dynamic coarse gestures} : this category is characterized by a single global trajectory of the hand without any movements of the fingers, and it includes 5 gesture classes : Left, Right, Circle, V and Cross.
    \item {\bf dynamic fine gestures} : this kind of gesture includes detailed movements of the fingers of the hand. This category includes 6 gesture classes  Grab, Pinch, Tap, Deny, Knob and Expand.
\end{itemize}
We introduce a "No gesture" label in order to distinguish non-significant sequences containing non-meaningful movements.

\subsection{Data preprocessing}
We use the same data preprocessing techniques that we used with the offline recognition method, translation and scale invariance techniques  because with most online gesture recognition datasets, the data is usually raw and unprocessed.
%\paragraph{Data augmentation}
We performed 3 augmentation steps on the training set, we use the same techniques applied in the offline HGR step: random moving and noise augmentation. \\
We use a new augmentation technique that we call {\bf sliding-window augmentation}, which consists in dividing the segmented gestures into small windows which would have the same label as their original gesture and augmenting the training set with these fragments. This augmentation technique would allow the model to learn small parts of the gesture and be able to classify a gesture before its end. This can be very useful in the online stage as the sequence window can contain only a small portion of the gesture and the rest would be labeled as "No gesture" and the model would be able to recognize the gesture from that small portion.
\subsection{Training and online evaluation}
Our STr-GCN model performs well and is quite efficient and the SHREC'21 dataset is very similar to the datasets used in the offline HGR, this should allow us to use STr-GCN in online HGR without changing the model architecture and get good results. In the rest of this section: we discuss the training process, we define our online evaluation technique and metrics, and we compare our results with the state-of-the-art.
\subsubsection{Training process}
The SHREC'21 dataset is an online hand gesture dataset and unlike the benchmarks used for offline HGR, the gesture samples are not segmented. The provided sequences can contain 3, 4 or up to 5 gestures and can contain "No gesture" fragments. \\
In order to train the model, we have to identify the "No gesture" parts of the sequence and segment the gestures out to be used as our training samples. We also sample some sequence fragments labeled as "No gesture" which we use to allow the model to recognize "No gesture" movements in an effort to reduce the false positives in the online recognition stage. \\
We use the same hyperparameters, training protocol and optimization techniques used in the first part of the project.
\subsubsection{Evaluation}
\subsubsection*{Evaluation protocol}
The online recognition task is different than the offline recognition, as the gestures are not segmented. There are 2 approaches to this: The gestures start and end frames should be detected, extract the sequence and then perform the recognition or the gestures should be recognized continuously.\\
In our work, we use a sliding window approach. At time step t, we crop a small window of the sequence, covering the interval [t, t+{window size}]. We sample windows with a stride of 5, that means windows are sampled every 5 frames, and we obtain a per-frame labeling of the window. After the window is sampled, we perform the recognition on it. But, we can't simply assign the label corresponding to the maximum probability as the prediction for this window, as the window can contain frames that belong to a gesture and frames that belong to the "No gesture" class. \\
There's a strong class imbalance, as most of the extracted windows belong to the "No gesture" class and would be classified as false positives. We need to consider a way to reduce the false positives' ratio, so we introduce a prediction filtering system based on probability thresholds. \\
The windows that are predicted to belong to a gesture class $C$ with a probability score $P(C)<\alpha(C)$ are considered as false positives and are considered to belong to the class "No gesture".\\
We learn these probability thresholds during the validation phase. The threshold $\alpha(C)$ for gesture class $C$ is the average probability estimated by the classifier that a gesture of class $C$ actually belongs to that class. The probability is only considered when the classifier correctly predicts the class of the gesture, so false positives will not contribute to the average threshold score of the gesture. \\
The use of this thresholds filtering system exponentially reduces the number of false positives.
\subsubsection*{Evaluation metrics}
The evaluation of the online recognition task is intrinsically different from the offline recognition task, where the accuracy is sufficient because the datasets are well-balanced, and the accuracy can give a reliable measure of the model's performance.
With the online recognition task, there's a strong class imbalance between the "No gesture" class and all the other classes, as most windows would be labeled as "No gesture". \\
We follow the evaluation protocol proposed in the SHREC'21 challenge \cite{shrec21}, they proposed the evaluation on three metrics:
\begin{itemize}
    \item {\bf Detection rate} : otherwise known as the true-positive rate, is the percentage of gestures of each class that were correctly predicted when compared with the ground truth gestures. The detection rate can be formulated as follows:
        \begin{equation}
\begin{split}
    {Detection \ rate}= \dfrac{TP}{( TP+FN )}
\end{split}
\end{equation}
    \item {\bf False positive rate} : is the ratio of the number of false predictions that do not match the ground truth samples of a particular class divided by the total number of ground truth samples belonging to this class. The false positive rate can be formulated as follows:
    \begin{equation}
\begin{split}
    {FP \ rate}= \dfrac{FP}{( TP+FN )}
\end{split}
\end{equation}
    \item {\bf Jaccard index} : The Jaccard index represents the average of the overlap between the predicted label and the ground truth label for a particular sequence (s) and label (l) pair. The result is the percentage of correctly labeled frames in the sequence. The Jaccard index can be formulated as follows:
        \begin{equation}
\begin{split}
    Jaccard Index= \dfrac{GT\textsubscript{s,l} \cap P\textsubscript{s,l} }{GT\textsubscript{s,l} \cup P\textsubscript{s}}
\end{split}
\end{equation}
    \end{itemize}
\subsubsection*{Results}
We collect the results of the SHREC'21 contest \cite{shrec21}, 4 research groups have participated. Table \ref{SOTA-online} presents the best reported result for each group and the results found by other state-of-the-art methods.
\begin{table}[!htbp]
\centering
    \begin{tabular}{p{1.8cm}|p{1.8cm}|p{1.8cm}|p{1.8cm}}
    \hline 
      {\textbf{Method}} & {\textbf{Detection rate }} & {\textbf{ Jaccard Index }} & {\textbf{ False-positive rate}}     \\ 
         \hline
        Group 1 (Run 3) \cite{shrec21} & 0.729  & 0.603 & 0.257   \\
        Group 2 (Run 1) \cite{shrec21} & 0.486  & 0.277 & 0.927   \\
        Group 3 (Run 2) \cite{shrec21} & 0.757 & 0.619 & 0.340   \\ 
        Group 4 (Run 3) \cite{shrec21} & 0.899 & {\bf 0.853} & {\bf 0.066}   \\ 
        STRONGER \cite{emporio2021stronger} & 0.906 & 0.740 & 0.347   \\ 
        \hline
        
        {\bf Ours} & {\bf 0.916} &  {\bf 0.853} &  0.083  \\
         \hline
    \end{tabular}
    \caption{ Results and comparison of our method with other state-of-the-art methods on the SHREC'21 dataset.}
    \label{SOTA-online}
\end{table}

As seen in table \ref{SOTA-online}, our method achieves good results, the second best results overall, but the false positive rate is slightly higher than the result of the best performing group. Although, group 4 used a gesture energy detection module that computes the shift of energy in each joint of the hand over a sequence of consecutive frames. They only perform the gesture recognition if an important energy shift is detected, otherwise the window is discarded. This explains the difference in results compared to the other groups and why their false positive rate is slightly lower.

\section{Conclusion}
In this work, we explored the problem of hand gesture recognition using 3D skeleton-based approaches. We reviewed the main challenges associated with offline and online recognition and analyzed existing state-of-the-art methods, particularly those based on GCNs, attention mechanisms, and transformers. Our proposed approach demonstrated competitive performance on three offline benchmarks—SHREC'17 Track, Briareo, and First Person Hand Action datasets—where it achieved results comparable to or surpassing state-of-the-art methods. Furthermore, we successfully adapted our model for online gesture recognition and evaluated it on the SHREC'21 dataset, highlighting its effectiveness in real-time applications.

For future work, we aim to integrate our hand gesture recognition system into human-robot interaction applications. Gesture-based communication can be used for direct robot control, enabling intuitive and efficient command execution, as well as for collaborative tasks where the robot adapts its behavior based on human gestures. This would open new possibilities in industrial environments, assistive robotics, and teleoperation scenarios.

Moreover, we plan to explore the use of additional modalities such as EEG to complement skeleton-based recognition. EEG signals can provide valuable information about user intentions, allowing for a more seamless and natural interaction, especially in scenarios requiring greater freedom of movement or where hand occlusion is an issue. Combining EEG and skeletal data could improve system robustness and open up new possibilities for multimodal gesture recognition in complex environments.


%% by copying the reference list from your .bbl file, paste it into the main manusctript .tex %%

%\bibliographystyle{unsrt}
%\bibliography{sn-article}% common bib file


%\bibliographystyle{unsrt} % Or IEEEtran, apalike, etc.
\bibliography{sn-article} % Ensure your .bib file is named correctly

\end{document}